# Evaluating Novel Mask-RCNN Architectures for Ear Mask Segmentation

Comparing Mask-RCNN variants across disparate datasets for Ear Mask segmentation

Saurav K. Aryal*

Electrical Engineering and Computer Science, Howard University; saurav.aryal@howard.edu

Teanna Barrett

Electrical Engineering and Computer Science, Howard University; teanna.barrett@bison.howard.edu

Gloria Washington

Electrical Engineering and Computer Science, Howard University; gloria.washington@howard.edu

The human ear is generally universal, collectible, distinct, and permanent. Ear-based biometric recognition is a niche and recent approach that is being explored. For any ear-based biometric algorithm to perform well, ear detection and segmentation need to be accurately performed. While significant work has been done in existing literature for bounding boxes, a lack of approaches output a segmentation mask for ears. This paper trains and compares three newer models to the state-of-the-art MaskRCNN (ResNet 101 +FPN) model across four different datasets. The Average Precision (AP) scores reported show that the newer models outperform the state-of-the-art but no one model performs the best over multiple datasets.

## 1 INTRODUCTION

Any human characteristic can be considered a biometric so long as it possesses the following qualities [10]:

- Universality: Everyone should possess said biometric.
- Uniqueness: No two persons should share the same identifying characteristic.
- Permanence: The identifying properties of the biometric need to be constant and unchanging over time.

Saurav K. Aryal, PhD and Gloria Washington, PhD are full-time faculty members of the Department of Electrical Engineering and Computer Science at Howard University. Teanna Barrett is a fourth-year bachelor's student of computer science at Howard University.

- Collectability: The characteristics need to be quantifiable and measurable.

While ears do fulfill these criteria, the ear has other features that make it appealing. These characteristics include, but are not limited to, the ability to capture images covertly from a distance, the capability of identifying monozygotic twins [23], and the possibility of supplementing other biometrics in multimodal systems [3]. Despite these valuable features, any system that utilizes ear-based recognition methods is bottlenecked by the underlying ear detection and segmentation approach. There is work that evaluates the performance of ear detection and segmentation for older techniques such as bounding boxes. These works contribute to the effective improvement of ear segmentation by identifying limitations such as poor performance in unconstrained image environments. Given that there are less reported evaluations of the state-of-the-art ear segmentation technique of ear masks, there is less understanding of the limitations of the more recent technique.

In this work, we hope to expand the existing literature on ear-mask segmentation for constrained and unconstrained datasets to provide insight into potential limitations. We train and evaluate the performance of Mask R-CNN ResNext-101 [18], BMask R-CNN ResNet-101 [7], and Cascade BMask R-CNN ResNet-101 [8] models and compare it to the current state-of-the-art Mask R-CNN model [5]. For this purpose, we train each model on four different datasets and report AP scores. The rest of the paper covers a review of related works in Section 2. The models and datasets utilized are detailed in Section 3. The training protocol and performance metrics are compared to the baseline model in Section 4. The limitations of the work are discussed in Section 5 and the final section outlines the conclusion and future work.

## 2 RELATED WORKS

The seminal work in ear detection and segmentation started with using the Hough Transform [2]. Since then, significant work has been done for ear localization and segmentation. These approaches can be broadly classified into Computer-Assisted ear segmentation, Template matching techniques, Shape-based Techniques, Morphological operators, Hybrid Methods, Harr-based Learning, and Deep Learning Based approaches [16]. It is worth noting that most of these approaches utilize bounding box detections. We strongly urge the reader to peruse existing survey papers on this topic to get a complete picture of existing approaches [16, 27, 29]. For the remainder of this section, we will review recent and best-performing approaches for ear segmentation. We cover bounding box detections and mask segmentation in two separate subsections.

### 2.1 Bounding Box Detections

An entropy-cum Hough-transform-based ear detection approach that employs an aggregate of hybrid ear localizer and an ellipsoid ear classifier to improve segment predictions was proposed [8]. Ear localization is considered correct if the detected area fully covers the ear and if the distance between the detection and target annotation centers is reasonable. The percentage of correct localization was 100.0% on both UMIST [17] and FEI [34]. On FERET [28], the score was lower at 73.95%.

Authors of [32] proposed a modified Hausdorff distance. Empirical results attest that the suggested method is invariant to shape, pose, illumination, and occlusion. The proposed technique was tested on both CVL face [26] and the ND-Collection E [25]. Detection rates of 91.0% and 94.5% were reported on the respective data sets.

An improved traditional Faster Region-based CNN algorithm with Multi-Scale Faster R-CNN framework was proposed in [35]. Their approach was evaluated on three different databases: UBEAR [26], AWE [14, 11, 12], and UND-J2 [36]. The authors achieved 100% accuracy on the constrained UND-J2 dataset. On the unconstrained, web-scraped AWE dataset, a 98% accuracy was achieved. Likewise, an accuracy of 89.66% was obtained on the grayscale UBEAR dataset. While the state of the art of bounding-box detections is impressive, there are a few caveats with any bounding box based approach:

- Since ears are not perfectly rectangular; even a perfect bounding box contains noise which may hamper

downstream recognition algorithms. Ear recognition algorithms are already susceptible to occlusion from hair and ear accessories. [13].
- While the issues outlined above can be resolved by adding a cleaning and extraction phase, which removes noise, it adds complexity and may introduce errors. We believe a single-stage instance mask segmentation approach is preferable to the standard bounding box or multi-step approaches for the aforementioned reasons.

### 2.2 Instance Mask Segmentation

Deep learning approaches for ear segmentation are fairly new. In [11] proposed a novel ear detection technique based on convolutional encoder-decoder model. On the AWE dataset, an average accuracy of 99.4% and a mean intersection-over-union (IoU) score of 55.7% was achieved. Recently, in 2019, the same group from [11] utilized a Mask-RCNN with a ResNet + FPN architecture [5] to improve upon their encoder-decoder model. This approach scored a mean IoU of 79.2%. While there have been improvements proposed, these approaches are either multi-step algorithms relying on face recognition models [24], require per-knowledge about the context of the ear image [15], or have not been tested on diverse skin tones [33]. As such, for our task [5] is the current state-of-art for ear mask segmentation and will be the baseline for performance comparison.

## 3 METHODOLOGY

This section details the models, datasets, and experimental procedure used in this work.

### 3.1 Models

Mask R-CNN [18] is simply an extension of the Faster R-CNN approach [29]. While Faster R-CNN performs bounding box detection, Mask R-CNN was implemented with the addition of a function for the prediction of an instance mask. The addition of mask predictions provides more detailed spatial information about the detected object by not compressing the feature information into vectors (as in the bounding box features). This added functionality is a fully connected convolutional network. When each region of interest (RoI) is passed through the post-processing branch, it returns a mask at the pixel-level for each instance in addition to the bounding boxes. The authors from [5], utilized this model with a ResNet101 + FPN open-sourced implementation from [1]. Henceforth, this model will be referred to as the baseline model.

The ResNeXt backbone provides performance improvements over the standard ResNet [19]. As such, for our first model, we chose to train a Mask R-CNN model with ResNeXt-101 + FPN in favor of the ResNet-101 used in the baseline. The second model, known as Boundary-preserving Mask R-CNN (BMask R-CNN) [7], is based on improvements to Mask R-CNN. It contains an added boundary-preserving mask head where object boundary and mask are jointly learned with fusion blocks. Consequently, the mask prediction results are better adjusted with object outlines. As such, BMask R-CNN exceeds Mask R-CNN by a significant margin on the COCO dataset. [8]. We utilize BMask-RCNN with a ResNet-101 +FPN backbone. The third model is a variant of BMask R-CNN called Cascade BMask RCNN. It has demonstrated better performance on the COCO dataset than the aforementioned BMask R-CNN [8]. However, at the time of writing, these models have been published for just under a year. As such, they have yet to be applied for improving mask segmentation tasks. To the best of our knowledge, this is one of the first works to utilize these two models for a domain-specific use case.

For this paper, the four methods were implemented with minor adjustments to utilize the open-source models. Considering the generalized nature of the models they are primed for the images from the COCO images dataset. However, for our purposes the ear image databases were transformed to be represented in COCO-style. In addition, the open-source implementations were adjusted in comparison to implementations described in the associated papers. For the baseline method, [1] mentioned image resizing, on the-fly bounding box generation, and reduction in learning rate adjustments made to the model used in the paper. For the Mask R-CNN model with ResNeXt-101 + FPN method, we used the same Mask R-CNN architecture but utilized detectron2 [35] for our implementation. The BMask R- CNN and the Cascade BMask R-CNN methods were implemented utilizing the official repository [8].

While there are improved models for instance segmentation [20], these approaches have not been tested on a relatively low resource problem such as ear segmentation. Furthermore, these models require large compute power for both training and inference. Therefore, such models were excluded from our analysis.

**3.2 Datasets**

Each of the datasets below is trained on each of the four models above, and the Average Precision (Metrics) is reported. For this work, we chose the following datasets:

*3.2.1 UND-J2 [36]*

This dataset contains 1800 3D and 2D side-profile PPM images from 415 human subjects captured in a constrained laboratory with good lighting and minimal occlusion. No mask or annotations are provided. Samples images of this dataset can be seen in Figure 1.

*3.2.2 UBEAR [30]*

This dataset is the only grayscale dataset provided. It contains JPG images captured in uncontrolled environments and under unconstrained protocols with subjects on the move and without requiring them any particular care about occlusions of the ears and poses. Binary masks are provided for only a portion of the images. Samples images of this dataset can be seen in Figure 2.

*3.2.3 UTSB-III [4]*

This dataset contains BMP ear images of 60 volunteers and has more variation than [36]. Every subject is photographed in three different images: normal frontal image, frontal image with trivial angular rotation, and image under varying illumination. No ground truths ear masks are provided. Samples images of this dataset can be seen in Figure 3.

*3.2.4 AWE [12, 13,16]*

This dataset contains 1,000 images of 100 persons. Images are unconstrained and were collected from the web using a semi-automatic procedure and contain the mask annotations as PNG files. Samples images of this dataset can be seen in Figure 4.

While there are other datasets, we believe, with the four datasets mentioned above, we cover constrained and unconstrained conditions, angular rotations, color-spaces, diverse skin tones, and occlusions which are the main use cases for ear-based recognition.

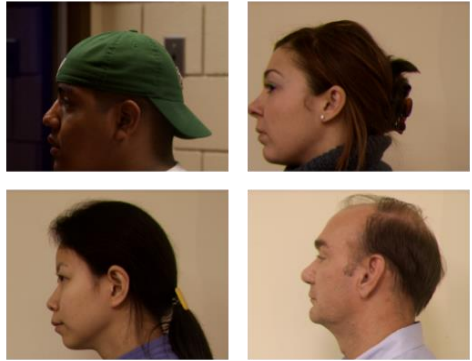

Figure 1: Sample Images from UND-J2 Collection Dataset

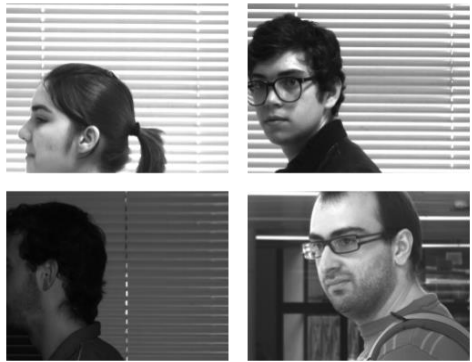

Figure 2: Sample Images from UBEAR Dataset

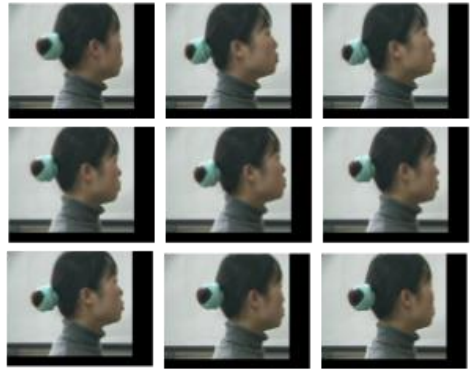

Figure 3: Sample Images from USTB-III Dataset

### 3.3 Experimental Procedures

This subsection will describe our data prepossessing, training-testing protocol, and evaluation metrics.

*3.3.1 Image Preprocessing*

Due to the ubiquity of bounding box approaches, the datasets, other than AWE and parts of UBEAR, did not come with ground truth mask annotations. As such, we had to carefully and manually label each image from UND-J2, USTB-III, and about half of the UBEAR datasets. This labeling was accomplished by using the open-source tool PixelAnnotationTool [6]. Ears were labeled as a gray masks with an RGB value of (35,35,35). Careful consideration was given to not label occlusions such as hair and accessories. A sample of the created mask and original image from the USTB-III dataset can be seen in Figure 5. When the corresponding masks were provided, they were formatted to match the manually labeled files' format and color-coding.

Since detectron2 does not explicitly support training with mask images, we also had to write a converter to convert mask images to the standard COCO JSON format [21]. For this purpose, based on our mask, we designated each pixel with an RGB value of (35,35,35) as Ear pixels. Every other color was considered a Not-Ear pixel. All datasets have input files of different formats. All images were converted to PNG format for the sake of convenience in training and testing. As the images from UBEAR were grayscale, they were converted to a 3 channel image by duplicating the pixel values across all channels.

*3.3.2 Modeling and Evaluation Metrics*

Each of the four datasets was trained on the four models outlined in the models section. We used a standard random 80/20 split for training and testing. The training and testing were performed on a Google Cloud Compute Engine instance with an NVIDIA-A100 GPU. For augmentation, we resized the shortest image and performed Random flips. To ensure fair comparison and considering available compute resources, we chose to train all models and datasets until convergence using batch size of 2, and a base learning rate of 0.01. No further hyper-parameter tuning was performed.

Performance metrics for ear segmentation tasks have been evolving inline with advancements in computer vision [16, 28, 29]. However, the current standard metric is still the mean IoU (Intersection over Union) score. Instead of existing metrics, which only given a mean measure of performance, we propose to utilize Average Precision (AP) metrics standardized for object segmentation task as part of the Common Objects in Context [22]. In particular, the pixel-level AP for each class: ear and not-ear is utilized. Also reported are the AP rates of ear segmentation task with an IoU=50 and IoU=75. Additionally, an 12-point interpolated average AP of IoU in the interval [0.50, 0.95] with 0.05 steps is evaluated for the entire test set to better measure overall segmentation performance.

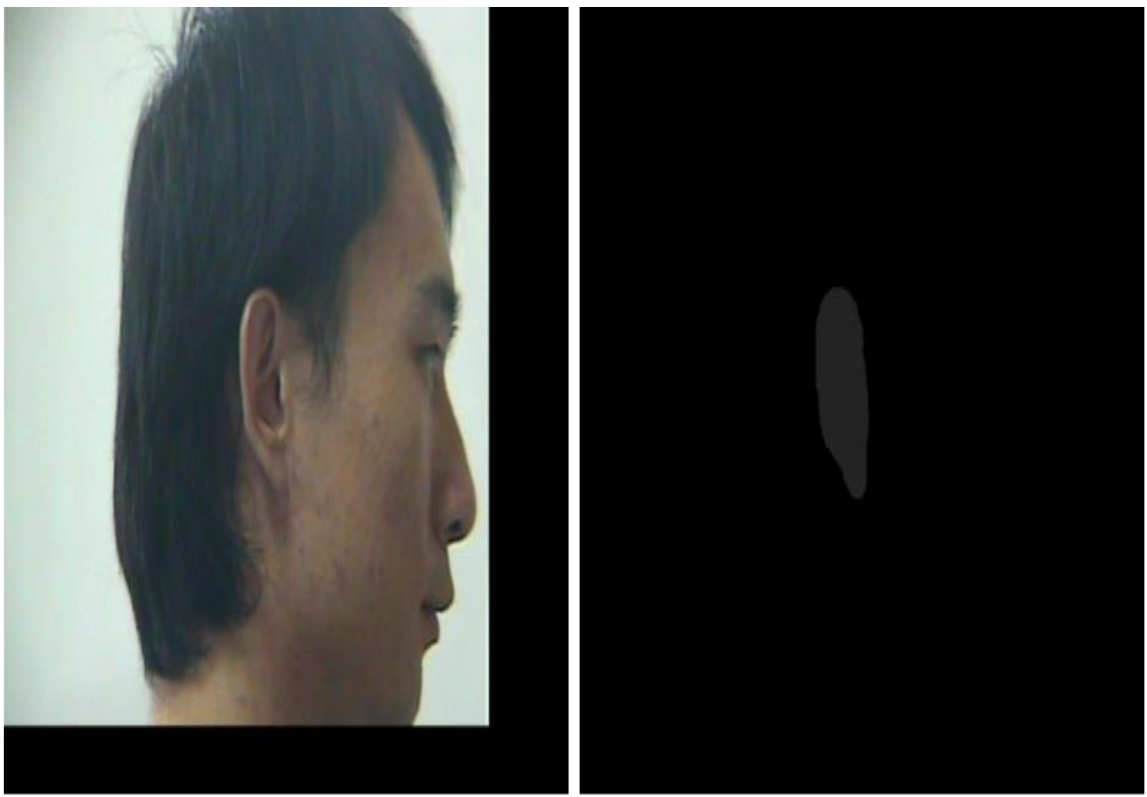

Figure 5: Example of original image and manually annotated mask image.

## 4 RESULTS

The following subsections describe the quantitative performance of each of the models on constrained (UND-J2), angular movements (USTB-II), unconstrained grayscale (UBEAR), and constrained color (AWE) environments. We then discuss sample qualitative results and discuss limitations and future work.

### 4.1 Quantitative Results by Dataset

#### 4.1.1 Constrained Laboratory Environment

The performance achieved on the UND-J2 dataset is indicative of the performance expected from these models in a real-world constrained environment in which there is very little variation in the positioning of subjects of users. The resulting AP scores for overall segmentation and Ear-vs-Not Ear segmentation can be seen in Table 1. The Mask R-CNN model with a ResNxt Backbone outperforms all other models and achieves 86.8% AP rate for ear segmentation. The baseline model is a close second. There very little performance difference in constrained environments across these models.

#### 4.1.2 Angular Movements

The performance achieved on the USTB-III dataset is indicative of the performance expected from these models in a real-world environment with angular movement, in which a subject may move in several positions during the recognition process. The resulting AP scores for overall segmentation and Ear-vs-Not Ear segmentation can be seen in Table 2. Similar to results in 4.1.1, the Mask R-CNN model with a ResNxt Backbone generally outperforms the other models. However, in this instance, the model clearly outperforms in all metrics with a slightly higher performance in ear segmentation than in a constrained environment (87.8% and 86.8% respectively).

*4.1.3 Grayscale and Unconstrained Environment*

The performance achieved on the UBEAR dataset is indicative of the performance expected from these models in a real-world unconstrained environment with a grayscale input. This type of environment can occur when a stream of images are collected by a grayscale camera or other grayscale imaging device. The resulting AP scores for overall segmentation and Ear-vs-Not Ear segmentation can be seen in Table 3. BMask R-CNN with a Resnet backbones does much better than the other models. Perhaps the loss of information with the loss of three channels was too much for most models. It is possible the boundary preserving behavior of BMask R-CNN helps mediate this loss of information. Despite the improvement in the BMask R-CNN with a Resnet backbone model, the AP scores are markedly lower than the previous environments. The current models may not be able to generalize as well to an unconstrained environment that has variations in image quality, occlusion presence and a variety of other inconsistencies.

*4.1.4 Color and Unconstrained Environment*

The performance achieved on the AWE dataset is indicative of the performance expected from these models in a real-world unconstrained environment. This environment is more accurate to how human subjects would position themselves and how their images would be captured for recognition. The resulting AP scores for overall segmentation and Ear-vs-Not Ear segmentation can be seen in Table 4. The Mask R-CNN model with a ResNext Backbone significantly outperforms all other models on all metrics except for Not-Ear segmentation. This environment has the lowest overall performance scores.

Table 1: Performance in Constrained Environment (UND-J2)

| Model | AP (Ear) | AP (Not Ear) | AP@0.50 | AP@0.75 | AP@0.50:0.95 |
|---|---|---|---|---|---|
| Baseline | 85.2 | 90.8 | **99.5** | **98.9** | 88.0 |
| Baseline + ResNxt | **86.8** | 91.3 | **99.5** | **98.9** | 89.0 |
| BMask R-CNN | 86.3 | **93.2** | 99.2 | 98.6 | **89.7** |
| Cascade BMask R-CNN | 84.3 | 92.0 | **99.5** | **98.9** | 88.2 |

Table 2: Performance with Angular Movements (USTB-III)

| Model | AP (Ear) | AP (Not Ear) | AP@0.50 | AP@0.75 | AP@0.50:0.95 |
|---|---|---|---|---|---|
| Baseline | 86.1 | 92.3 | **98.7** | **98.7** | 89.2 |
| Baseline + ResNxt | **87.8** | 93.8 | **98.7** | **98.7** | **90.8** |
| BMask R-CNN | 87.4 | 90.9 | 98.6 | 98.6 | 89.1 |
| Cascade BMask R-CNN | 87.5 | 93.7 | 98.6 | 98.6 | 90.6 |

Table 3: Performance in Grayscale and Unconstrained Environment (UBEAR)

| Model | AP (Ear) | AP (Not Ear) | AP@0.50 | AP@0.75 | AP@0.50:0.95 |
|---|---|---|---|---|---|
| Baseline | 76.2 | 89.9 | 98.5 | **94.3** | 83.0 |
| Baseline + ResNxt | 63.6 | 81.6 | 90.8 | 84.2 | 72.6 |
| BMask R-CNN | **77.6** | 88.8 | **98.9** | 92.6 | **83.2** |
| Cascade BMask R-CNN | 62.5 | **90.5** | 90.6 | 83.9 | 76.5 |

Table 4: Color and Unconstrained Environment

| Model | AP (Ear) | AP (Not Ear) | AP@0.50 | AP@0.75 | AP@0.50:0.95 |
|---|---|---|---|---|---|
| Baseline | 62.6 | 85.6 | 92.7 | 84.6 | 74.1 |
| Baseline + ResNxt | **69.8** | 85.7 | **97.4** | **87.2** | **77.8** |
| BMask R-CNN | 66.5 | 79.6 | 95.2 | 85.6 | 73.1 |
| Cascade BMask R-CNN | 64.7 | **86.7** | 93.4 | 85.8 | 75.9 |

### 4.2 Qualitative Analysis

For unconstrained scenarios, failed defections and false positives persist. Two examples of failed, partially correct (inaccurate), and near-perfect (best) detections are shown for the baseline, Mask-RCNN model and the best performing model for the UBEAR and AWE datasets, respectively.

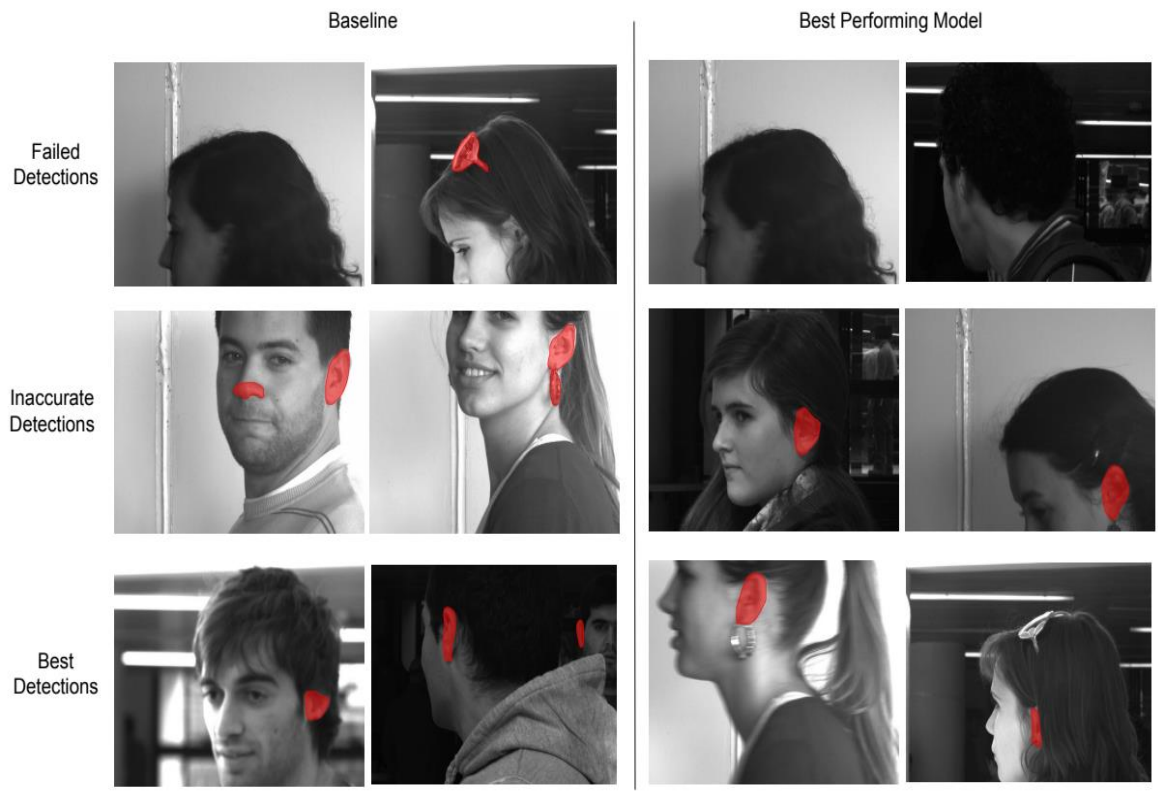

Figure 6: Samples of Ear Segmentation on the UBEAR dataset.

The quantitative performance metrics indicate that unconstrained environments tend to exhibit the worst performance from all selected models. In particular, we notice, similar to the baseline, results vary between extremely well segmented, partially accurate, and complete failures on both UBEAR and AWE datasets. Most prominent inaccurate segmentations revolve around incorrect segmentation of the nose and accessories such as earrings and glasses. It is possible that the model is learning there should be two ears should the image contain a fully visible face which seems to be the case for most inaccurate detections. On the other hand, we noticed multiple failed detections for high levels of occlusions and low lighting environments. Perhaps with larger datasets and more work on low resource models could bridge this current performance gap.

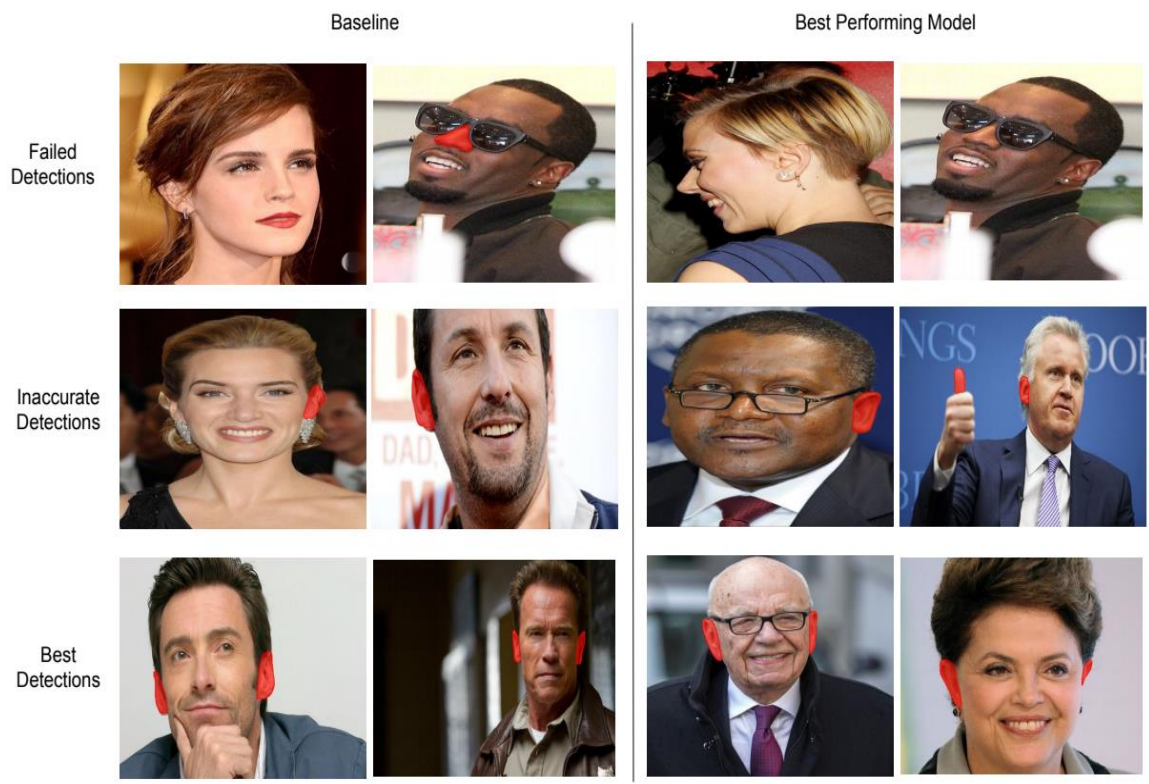

Figure 7: Samples of Ear Segmentation on the AWE dataset.

## 5 DISCUSSION

The major limitations of our approach include the models used as well as data availability and heterogeneity. While newer variants of Mask RCNN were applied and increased performance on advanced metrics, no novel model was inherently proposed. Regardless, this work contributes to the niche field of ear mask segmentation and demonstrates the possibility of open sourced models to provide state-of-the-art results. However, analysis related to performance discrepancy resulting from occlusions, skin tones, and illuminations were not directly evaluated and should be considered future work. Since there is no one-model-fits-all solution for ear segmentation as of yet, this may also be a promising research direction. Furthermore, while the current approaches should theoretically extend to multiple subjects per image, current datasets focus on mostly one subject per image and lack subjects with darker skin tones. These data gaps are yet to be addressed and may be future work. All existing datasets also do not provide instance masks currently and expect researchers to create their own annotations. Nonetheless, more work needs to be done in creating instance segmentation datasets for ear biometrics and models that work across varying constraints.

## 6 CONCLUSION

In conclusion, we noticed that there is no one-model-fits-all solution for ear segmentation as of yet. However, for grayscale images, BMask-RCNN models perform much better than the rest in terms of AP scores for instance segmentation. In contrast, Mask R-CNN with a ResNext backbone does, on average, better than all other models across the multiple datasets used. While the current approaches should extend to multiple subjects per image, current datasets focus on mostly one subject per image. We also notice a lack of darker skin tones These data gaps are yet to be addressed and may be future work. Although we are closer to realizing real-world ear segmentation, more work needs to be done in creating instance segmentation datasets for ear biometrics and models that work across varying constraints.

## ACKNOWLEDGMENTS

This project was supported (in part) by the National Security Agency, USA (H98230-19-P-1674), Northrop Grumman (#201603), National Science Foundation, USA (#1828429), and University of Maryland Applied Research Laboratory for Intelligence and Security (#1828429). Additionally, the computational resources were provided through a Google Cloud Research Credit Grant. The content is solely the responsibility of the authors and does not necessarily represent the official views of the funding organizations.